\documentclass[10pt,journal,compsoc]{IEEEtran}
\ifCLASSOPTIONcompsoc
  \usepackage[nocompress]{cite}
\else
  \usepackage{cite}
\fi

\usepackage{graphicx}
\usepackage{subfigure}
\usepackage{color}

\usepackage{booktabs}
\usepackage{bm}
\usepackage{amsfonts}
\usepackage{amsmath}
\usepackage{multirow}
\usepackage{url}

\ifCLASSINFOpdf
\else
\fi
\begin{document}
%
\title{Cross-spectral Face Completion for NIR-VIS Heterogeneous Face Recognition}
%
%
%
%

\author{Ran~He,~\IEEEmembership{Senior Member,~IEEE,}
        Jie~Cao, Lingxiao Song,
        Zhenan~Sun,~\IEEEmembership{Member,~IEEE,}
        and~Tieniu~Tan,~\IEEEmembership{Fellow,~IEEE}
\IEEEcompsocitemizethanks{\IEEEcompsocthanksitem
R. He, J. Cao, L. Song, Z. Sun and T. Tan are with National Laboratory of Pattern Recognition, CASIA,
Center for Research on Intelligent Perception and Computing, CASIA,
Center for Excellence in Brain Science and Intelligence Technology, CAS
and University of Chinese Academy of Sciences, Beijing, China,100190.\protect\\
E-mail: \{rhe, jie.cao, lingxiao.song, znsun, tnt\}@nlpr.ia.ac.cn
}
\thanks{Manuscript received Janurary xx, 2019; revised August xx, 2019.}}

%
%

\markboth{Journal of \LaTeX\ Class Files,~Vol.~xx, No.~x, Janurary~2019}%
{Shell \MakeLowercase{\textit{et al.}}: Bare Demo of IEEEtran.cls for Computer Society Journals}
%



\IEEEtitleabstractindextext{%
\begin{abstract}

Near infrared-visible (NIR-VIS) heterogeneous face recognition
refers to the process of matching NIR to VIS face images. Current
heterogeneous methods try to extend VIS face recognition methods to
the NIR spectrum by synthesizing VIS images from NIR images.
However, due to self-occlusion and sensing gap, NIR face images lose
some visible lighting contents so that they are always incomplete
compared to VIS face images. This paper models high resolution
heterogeneous face synthesis as a complementary combination of two
components, a texture inpainting component and pose correction
component. The inpainting component synthesizes and inpaints VIS
image textures from NIR image textures. The correction component
maps any pose in NIR images to a frontal pose in VIS images,
resulting in paired NIR and VIS textures. A warping procedure is
developed to integrate the two components into an end-to-end deep
network. A fine-grained discriminator and a wavelet-based
discriminator are designed to supervise intra-class variance and
visual quality respectively. One UV loss, two adversarial losses and
one pixel loss are imposed to ensure synthesis results. We
demonstrate that by attaching the correction component, we can
simplify heterogeneous face synthesis from one-to-many unpaired
image translation to one-to-one paired image translation, and
minimize spectral and pose discrepancy during heterogeneous
recognition. Extensive experimental results show that our network
not only generates high-resolution VIS face images and but also
facilitates the accuracy improvement of heterogeneous face
recognition.
\end{abstract}

}

\maketitle

\IEEEdisplaynontitleabstractindextext

%
\IEEEpeerreviewmaketitle

\IEEEraisesectionheading{\section{Introduction}\label{sec:intro}}
\IEEEPARstart{I}{llumination} variation is a traditional challenge
in real-world face recognition systems. Near infrared (NIR) imaging
provides a low-cost and effective solution to acquire high-quality
images in low lighting or complete darkness conditions. Hence, it
has been widely adopted in mobile device, video surveillance and
user authentication applications. However, many applications require
that the enrollment of face templates is based on visible (VIS)
images, such as online registration and pre-enrollment using
passport or ID card. That is, NIR images are face images captured
under near infrared lighting, and VIS images are face images
captured under visible lighting. Therefore, face matching between
NIR and VIS images has drawn much attention in computer vision and
machine learning. It also has been the most studied research topic
in heterogeneous face recognition (HFR) that refers to matching
faces across different spectral (or sensing) domains and is
different from conventional VIS face recognition under homogeneous
conditions~\cite{SOuyang:2016}\cite{RHe:2018}.

\begin{figure}[t]
\begin{center}
\includegraphics[width=0.9\linewidth]{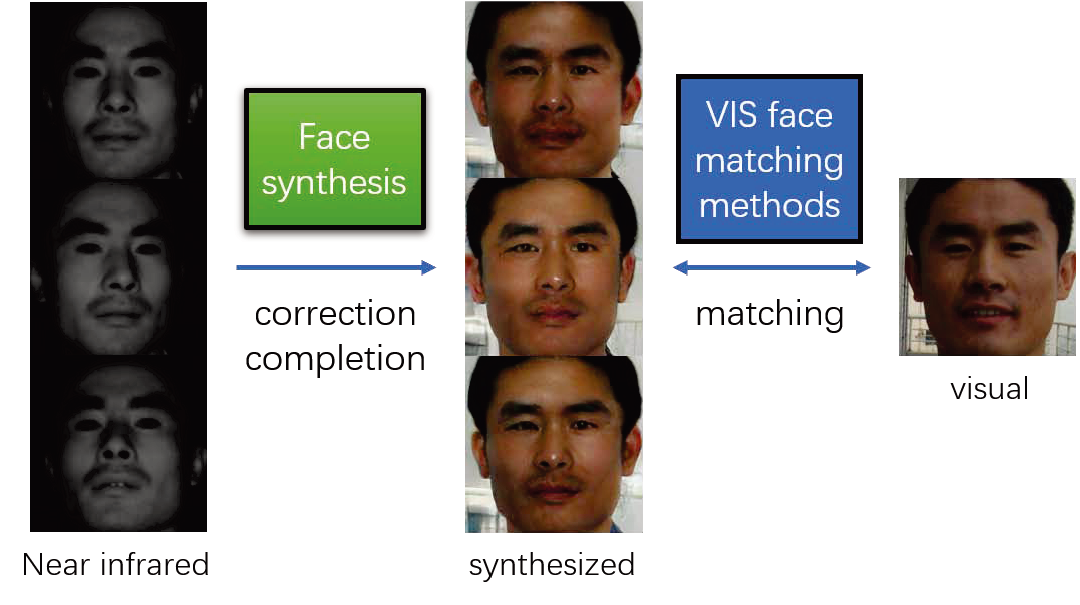}
\end{center}
   \caption{Synthesizing visible faces from near infrared faces is an unsupervised image translation problem because there is no exact pixel-level correspondence between the images from different spectral domains. Self-occlusion and sensing gap make some pixels or contents of a near infrared face occluded or corrupted.
   Given one near infrared face image as the input, cross-spectral face completion can produce a high-resolution and frontal visible face image. \label{fig:problem}}
\end{figure}

Since it is expensive and time-consuming to obtain a large-scale
pair-wised face images from different domains, current deep HFR
methods mainly resort to the convolutional neural network (CNN)
trained on a web-scale VIS face dataset, and then fine-tune it on a
NIR-VIS dataset to obtain better HFR
performance~\cite{Reale:2016}\cite{RHe:2017}\cite{Sarfraz:2017}.
Recently, to extend VIS face recognition methods to other spectral
domains, face synthesis methods have gained much
attention~\cite{Riggan:2016}\cite{HZhang:2017}\cite{Lezama:2017}\cite{LXSong:2018}.
Lezama et al.~\cite{Lezama:2017} proposed a cross-spectral
hallucination and low-rank embedding to synthesize a heterogeneous
image in a patch way. Song et al.~\cite{LXSong:2018} employed
generative adversarial networks (GAN)~\cite{Goodfellow:2014} with a
two-path model to synthesize VIS images from NIR images.
\cite{Riggan:2016}\cite{HZhang:2017} synthesize visible face images
from thermal face images. One major advantage of these synthesis
methods is that given the synthesized visible face images, any VIS
face recognition method trained on VIS face data can be used to
match the synthesized image to the enrolled VIS
images~\cite{Lezama:2017}\cite{HZhang:2017}.

The use of these synthesis methods poses opportunities as well as
new challenges. Current synthesis results are less appealing in
high-resolution and their output size is often no larger than $128
\times
128$~\cite{Riggan:2016}\cite{HZhang:2017}\cite{Lezama:2017}\cite{LXSong:2018}.
One possible reason lies in sensing gap. That is, VIS and NIR face
images of the same subject are captured by using different sensory
devices with different settings so that their visual appearances are
significantly different. Both geometric and textural details of NIR
faces are different from those of VIS faces. This gap results in
high intra-class variations and makes the high-resolution synthesis
of one spectrum from another very difficult. Particularly, as shown
in Fig.~\ref{fig:problem}, some visual appearances or contents are
often always missed or corrupted in NIR images (e.g., the pixels
around cheek, hair or eyes) so that the synthesis of visible images
is also a challenging inpainting problem.

Another possible reason making the synthesis challenging lies in the
pose difference between VIS faces and NIR faces~\cite{RHe:2018}.
Pose variations often result in self-occlusion so that the texture
of a NIR face image may be incomplete~\cite{JDeng:2018}. Since VIS
faces and NIR faces are often captured under different distances and
environments, it is difficult to simultaneously capture the VIS
faces and NIR faces under the same pose. For some real-world
applications on mobile devices, there are often various poses in NIR
face images~\cite{SLi:2013}. On contrast, poses of the pre-enrolled
VIS images using passport or ID card are often frontal. Moreover,
compared to the dataset for the synthesis in VIS domain, the NIR-VIS
dataset is often small-scale. Since a small-scale training dataset
will lead to the over-fitting problem~\cite{RHe:2018},
cross-spectral face rotation is more challenging than face rotation
in VIS domain. Both sensing gap and pose difference make the
synthesis from NIR to VIS be an one-to-many unsupervised
image-to-image translation problem as shown in
Fig.~\ref{fig:problem}.




To address the above two issues, this paper proposes an end-to-end
generative framework, named Cross-spectral Face Completion (CFC), by
performing generative adversarial networks. CFC presents a deep
framework for generating a frontal VIS image of a person's face
given an input NIR face image. It decomposes the unsupervised
heterogeneous synthesis problem into two complementary problems,
generating a texture inpainting component and a pose correction
component that are addressed by deep networks. The inpainting
component synthesizes and inpaints VIS image textures from
incomplete NIR image textures. The correction component transforms
any pose in NIR images to a frontal pose in VIS images. Once the
face pose of a VIS face image is given, the texture synthesis and
inpainting become one-to-one supervised image-to-image translation
problem, which results in paired NIR and VIS textures and
facilitates pixel-level losses. Second, a warping procedure is
developed to integrate the texture and pose procedures into an
end-to-end deep network. A fine-grained discriminator is employed to
supervise the disentanglement process by guiding the generator to
minimize the intra-class variance in an adversarial manner; a
wavelet-based discriminator is designed to supervise the visual
quality in a multi-scale manner. One UV loss, two adversarial losses
and one pixel loss are imposed to ensure high-quality synthesis
results.


We train our proposed CFC approach only on CASIA NIR-VIS
2.0~\cite{SLi:2013}, and evaluate it on CASIA NIR-VIS 2.0,
BUAA-VisNir~\cite{DHuang:2012}, and Oulu-CASIA~\cite{JChen:2009}. A
new benchmark 'recognition via generation' protocol on CASIA NIR-VIS
2.0 is established for systematically evaluating NIR-VIS
cross-spectral synthesis. Experimental results verify that by
attaching the pose correction procedure, we can simplify
cross-spectral face completion from one-to-many unpaired image
translation to one-to-one paired image translation, and minimize
spectral and pose discrepancy during heterogeneous recognition.
Extensive cross-database experiments show that our approach not only
generates photo-realistic and identity-preserving VIS face images
and but also facilitates HFR performance.


In summary, the main contributions of this work are as follows,
\begin{itemize}\setlength{\itemsep}{1pt}
\item A novel GAN-based end-to-end deep framework is proposed for cross-spectral face synthesis without assembling multiple image patches.
It contains encoder-decoder structured generators and two novel
discriminators to fully consider variations of NIR and VIS images.
\item It is the first time that the unsupervised heterogeneous face synthesis
problem is simplified to a one-to-one image translation problem. The
decomposition of texture inpainting and pose correction enables the
generation of realistic identity preserving VIS face images
possible.
\item This is the first approach simultaneously transforming face pose
and cross-spectrum appearance for heterogeneous face recognition. A
new benchmark on CASIA NIR-VIS 2.0 is also established to
quantitatively evaluate the performance of
'recognition via generation'. 
\item We achieve state-of-the-art face synthesis and face recognition
performance on multiple HFR benchmark datasets, including CASIA
NIR-VIS 2.0, BUAA-VisNir, and Oulu-CASIA. We synthesize $256 \times
256$ visible faces to push forward the advance in cross-spectral
face synthesis.
\end{itemize}

An early version of this work was first proposed
in~\cite{LXSong:2018}. Although \cite{LXSong:2018} and this work are
both based on GANs, they adopt different strategies to address the
unsupervised heterogeneous face synthesis problem.
\cite{LXSong:2018} employed the cycle-GAN
architecture~\cite{JZhu:2017} to handle the unsupervised synthesis
problem and a two-path network structure to enhance local textures.
On contrast, this paper simplifies the unsupervised synthesis
problem to a supervised one by decomposing the synthesis into two
complementary components. This decomposition has significantly
extended our previous work \cite{LXSong:2018} in network structure
and loss function, resulting in a high-resolution synthesis image.
Particularly, different from \cite{LXSong:2018} that trained a
feature representation on synthesized VIS images, this paper
directly uses synthesized visible images for HFR without fine-tuning
VIS face recognition models on synthesized images.

\section{Background and Related Work \label{sec:rw}}
The heterogeneous problem of matching people across different
domains has received increasing attention in biometrics (e.g.,
face~\cite{SLi:2006} and iris~\cite{LXiao:2013}). NIR-VIS HFR has
been one of the most extensively studied topics in heterogeneous
biometrics. In this section, we mainly review some recent advances
related to the heterogeneous matching problem from three
aspects~\cite{JZhu:2014}\cite{SOuyang:2016}: image synthesis, latent
subspace, and domain-invariant features.

{\bf Image synthesis} methods aim to synthesize face images from one
domain into another so that heterogeneous images can be directly
compared in the same spectral domain. These methods bridge the
domain discrepancy at the image preprocessing stage. They transform
face images from one domain to another, and thereby perform face
matching~\cite{RWang:2009}. Image synthesis was firstly studied to
synthesize and recognize a sketch image from a face
photo~\cite{XTang:2003}. \cite{RWang:2009} proposed an
analysis-by-synthesis framework to synthesize a face image from one
domain to another before face matching. \cite{XWang:2009} applied
Markov random fields to transform pseudo-sketch to face photo in a
multi-scale way. \cite{ZLei:2008r} resorted to statistical learning
to synthesize a 3D face from a single NIR face image using canonical
correlation analysis (CCA). In
\cite{SWang:2012}\cite{DHuang:2013}\cite{FXu:2015}, coupled or joint
dictionary learning was used to reconstruct face images and then
perform face matching. These three methods constrain the
representation of heterogeneous images in each dictionary to be the
same. \cite{MZhang:2019} novelly decomposed sketch-photo face
synthesis into an inter-domain transfer process and an intra-domain
transfer process and hence proposed a dual-transfer synthesis
framework.

Based on the recent advances in deep learning, \cite{Riggan:2016}
proposed a two-step procedure (VIS feature estimation and VIS image
reconstruction) to synthesize VIS faces from polarimetric thermal
faces. \cite{HZhang:2017} developed a fusion technique to
concatenate different stokes images for VIS face synthesis.
\cite{Lezama:2017} proposed a cross-spectral synthesis and low-rank
embedding to synthesize a heterogeneous image in a patch way. To
achieve better rank-1 accuracy, \cite{Lezama:2017} used a new
testing protocol rather than the standard 10-fold testing
protocol~\cite{SLi:2013}. \cite{RHuang:2017} proposed a global and
local perception GAN to reduce the pose discrepancy between a face
profile and a frontal face. Inspired by \cite{RHuang:2017},
\cite{LXSong:2018}~employed generative adversarial networks to
perform cross-spectral image synthesis. A two-path model was
introduced to alleviate the lack of paired images. Although deep
learning methods have significantly improved synthesis results,
synthesizing a heterogeneous image from another domain is still
challenging and the output size is often no larger than $128 \times
128$. As shown in Fig.~\ref{fig:compare_res} and
Fig.~\ref{fig:visual_res}, this synthesis is also an inpainting or
completion problem because background contents of NIR images are
corrupted during sensing.

{\bf Latent subspace} methods project two different domains to a
common latent space, in which the relevance of heterogeneous data
can be measured. Dimension reduction techniques such as Principal
Component Analysis (PCA), Canonical Correlation Analysis (CCA) and
Partial Least Squares (PLS) are often used. \cite{DLin:2006}
proposed a method called Common Discriminant Feature Extraction
(CDFE) to incorporate both discriminative and locality information.
\cite{ZLei:2012} considered the locality information in kernel space
and proposed a coupled discriminant analysis. Then,
\cite{XHuang:2013} developed a regularized discriminative spectral
regression method to seek a common spectral space. A common subspace
learning method was proposed in \cite{KWang:2016} by introducing
feature selection, and then was applied as a baseline method in
NIR-VIS HFR. \cite{Klare:2013} proposed a prototype random subspace
method with kernel similarities for HFR. \cite{CHou:2014} proposed a
domain adaptive self-taught learning approach to derive a common
subspace. By using transfer learning, \cite {MShao:2014} projected
both NIR and VIS data to a generalized subspace where each NIR
sample can be represented by some combination of VIS samples.
\cite{DYi:2015} employed Restricted Boltzmann Machines (RBMs) to
learn a shared representation between different domains, and then
suggested to apply PCA to remove the redundancy and heterogeneity.
Multi-view discriminant analysis~\cite{MKan:2016} and mutual
component analysis~\cite{ZFLi:2016} were further developed to reduce
the domain discrepancy. \cite{YJin:2017} integrated multi-task
clustering with extreme learning machine to learn coupled mapping
for NIR-VIS HFR. \cite{JGui:2018} treated HFR as a multi-view
discriminant analysis problem and projected the examples from
different modalities (or views) to one discriminant common space.


{\bf Domain-invariant feature} methods seek discriminative features
that are only related to face identity and disregard domain
information. Traditional methods in this category are almost based
on handcrafted local features, such as Local Binary Patterns (LBP),
Histograms of Oriented Gradients (HOG) and Difference of Gaussian
(DoG)
\cite{SLiao:2009}\cite{Klare:2011}\cite{Goswami:2011}\cite{Dhamecha:2014}.
To capture high-level semantics across different domains,
\cite{MShao:2016} constructed a hierarchical hyper lingual-words
based on bag of visual words. \cite{DGong:2017} converted facial
pixels into an encoded face space with a trained common encoding
model. Recently, deep neural networks show great potential to learn
domain-invariant features of heterogeneous images. These deep
methods are often pre-trained on a large-scale VIS face dataset, and
then are fine-tuned on NIR face images to learn domain-invariant
features.

Based on a pre-trained VIS CNN, \cite{Saxena:2016} explored
different metric learning strategies to improve HFR performance.
\cite{XXLiu:2016} designed two types of NIR-VIS triplet loss to
reduce the domain discrepancy meanwhile to augment training sample
pairs. \cite{Reale:2016} gave two new network structures (named
VisNet and NIRNet) with small convolutional filters, and used a
Siamese network to couple the learnt features from the two networks.
\cite{Sarfraz:2017} employed deep neural networks to learn a
non-linear mapping to bridge the domain gap between thermal and VIS
face images. \cite{RHe:2017} divided the high-level layer of CNN
into two orthogonal subspaces so that domain-invariant identity
information and domain-related spectrum information can be presented
independently. \cite{XWu:2019} explored a disentangled latent
variable space to optimize the approximate posterior for NIR and VIS
features. \cite{RHe:2018} designed a Wasserstein CNN to reduce the
discrepancy between NIR and VIS feature distributions. By performing
deep features, these methods significantly improve HFR results
against traditional methods.

Although some HFR methods have been developed to improve recognition
performance, HRF is more challenging than VIS face recognition. The
high performance of VIS recognition benefits from deep learning
techniques and large amounts of VIS face images. Due to limited
samples and sensing gap, HFR is still a challenging recognition
research topic. Moreover, the pose difference between NIR and VIS
faces also affects HRF performance and draws less attentions. Deep
learning based image synthesis methods pose opportunities as well as
new challenges.

\begin{figure*}
\centering
\subfigure[]{\includegraphics[width=0.9\textwidth]{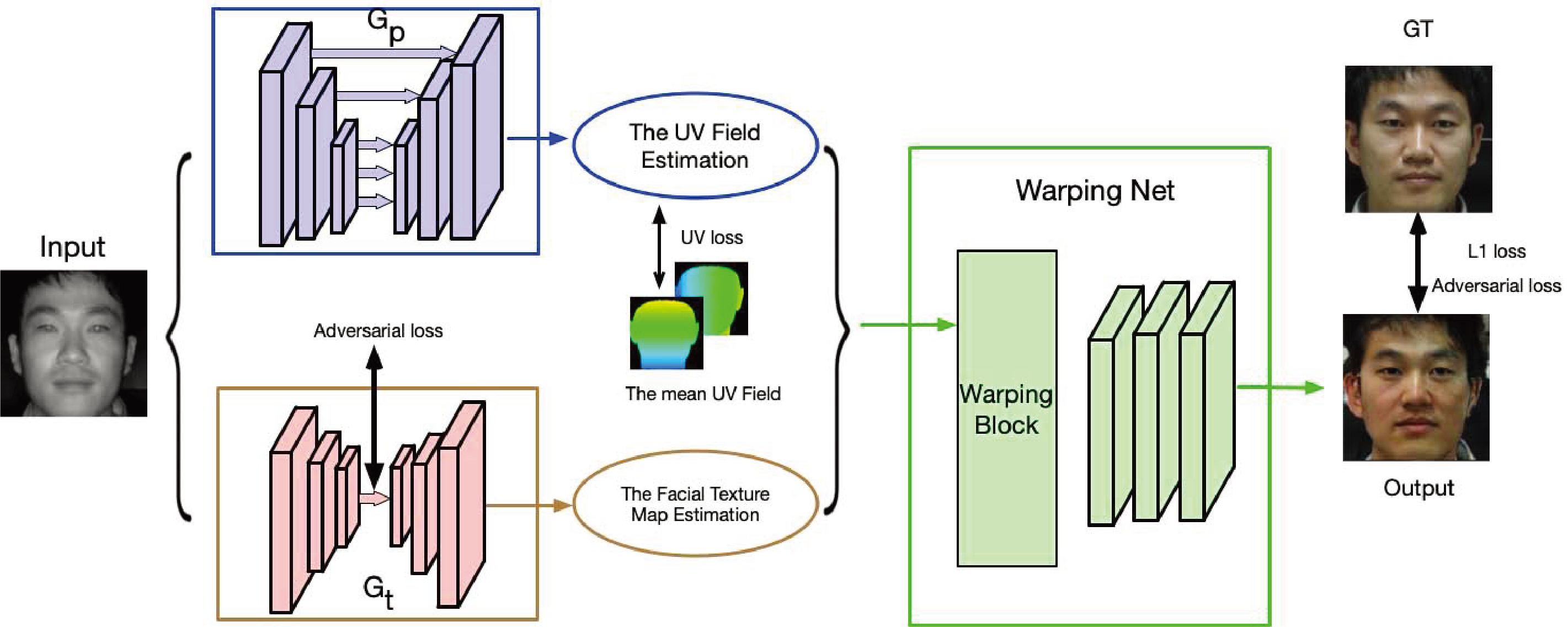}}
\subfigure[]{\includegraphics[width=0.42\textwidth]{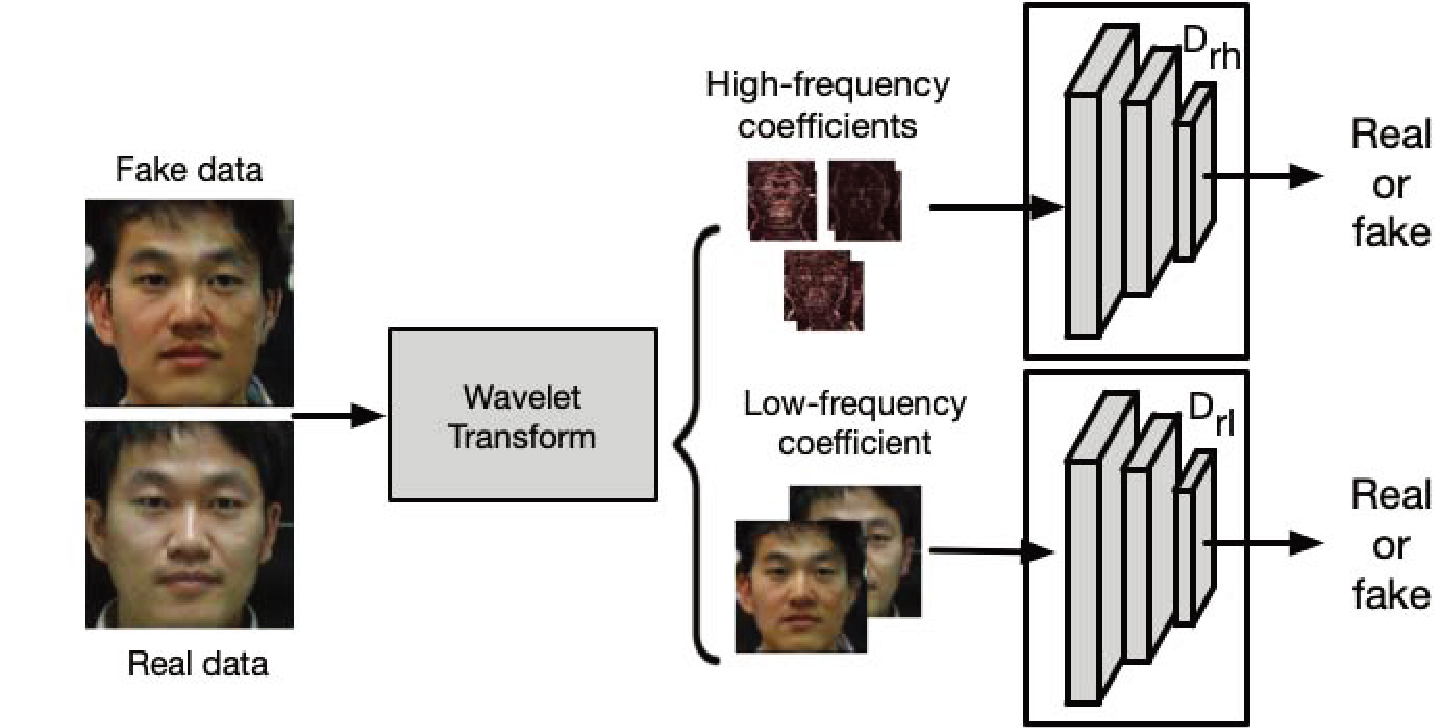}
\quad} \subfigure[]{\quad
\includegraphics[width=0.42\textwidth]{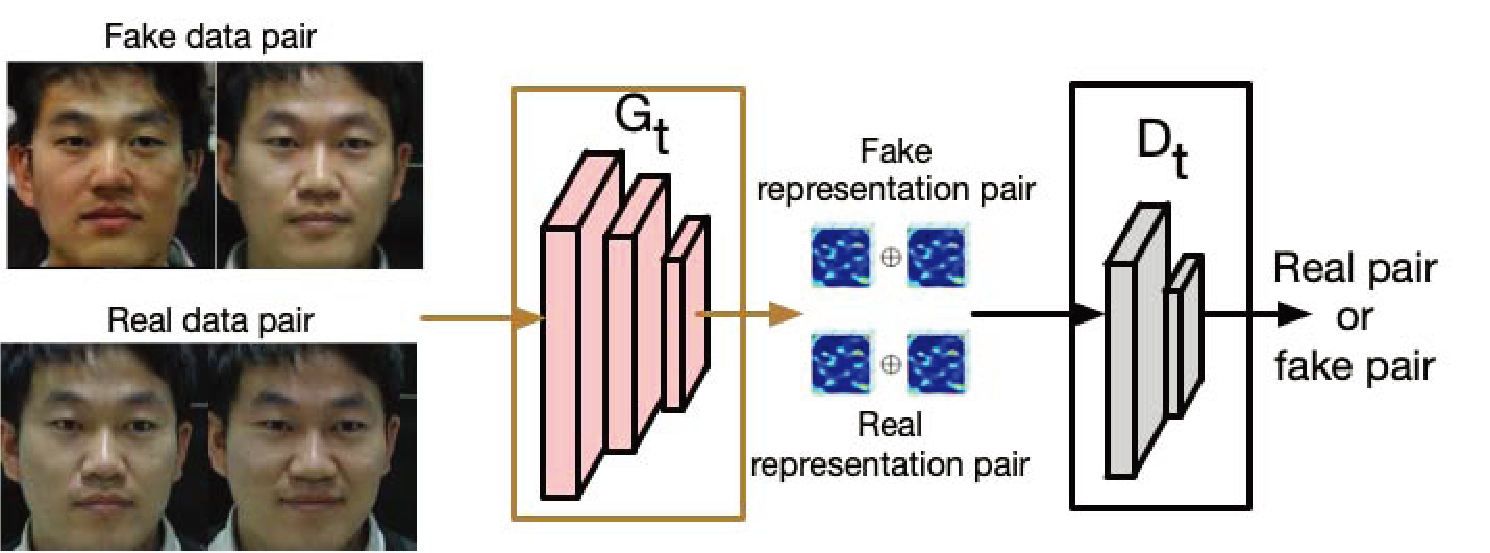}} \caption{An
illustration of our NIR-VIS face completion network. (a), (b), (c)
depict the of forward propagation processes of the generator, the
multi-scale discriminator, and the fine-grained discriminator,
respectively. \label{fig:illu}}
\end{figure*}

\section{Cross-spectral Face Completion}
The NIR-VIS face completion problem can be formulated as learning a
mapping from face images in a visual domain $X \in R^{H \times W
\times 3}$ to another visual domain $Y \in R^{H \times W \times 3}$.
For each face, there is an identity label whose corresponding
identity information should be well preserved in the completion
procedure. Based on practical applications, we assume that each
identity is included in both $X$ and $Y$ during the training
process. However, as we pointed out in Sec.~\ref{sec:intro}, data
captured by the devices without synchronization settings will
introduce large variations in pose, expressions, background, and
etc. Therefore, the training data are not simply regarded as
strictly paired. 

Our network is established in an adversarial learning framework as
shown in Fig.~\ref{fig:illu}. The generation modules consist of a
pose correction network $G_p$, a texture inpainting network $G_t$,
and a backend network named fusion warping net. Given an input face,
$G_p$ is trained to estimate the normalized shape information with
the aid of the dense UV correspondence field. $G_t$ aims to learn to
produce pose-invariant facial texture representation. The fusion
warping net combines the corrected shape and completed texture
information, and then produces the final results. There are two
discriminators in our network, a multi-scale discriminator $D_r$ and
a fine-grained discriminator $D_t$. The former is designed to
supervise the visual quality by discriminating between real images
in domain $Y$ and translated images. The latter aims to supervise
the disentanglement process by guiding the generator to minimize the
intra-class variance of the input in an adversarial manner. In the
following, we will describe the details of the pose correction,
texture inpainting, and fusion warping processes in
Sections~\ref{sec:pose}, \ref{sec:inpa}, and \ref{sec:fuse},
respectively. The overall loss function is summarized in
Sec.~\ref{sec:all}.

\subsection{Pose Correction via UV Field Estimation \label{sec:pose}}
The pose correction network $G_p$ is based on the estimation of the
dense $UV$ correspondence field that denoted as the $UV$ field in
the following part. The $UV$ field is employed to bind the UV facial
texture space and the RGB color image space. The $UV$ facial texture
space refers to the space where the manifold of the face is
flattened into a contiguous 2D atlas. Compared with shape guidance
information applied widely in face manipulation, e.g., landmarks and
facial parsing maps, the $UV$ field has two significant advantages:
1) the pixel-wise relation between the pose-invariant facial texture
map and the 2D image is specified by the UV field. Therefore, the
$UV$ field contains complete shape information for a given face. 2)
3D supervision is subtly integrated by the UV field. Since a facial
texture map in the $UV$ space represents the flattened surface of a
3D human face, the $UV$ field makes our approach 3D-aware.

To obtain the ground truth $UV$ field during the training process,
we fit the 3DMM (provided by \cite{Paysan:2009}) through the
Multi-Features Framework \cite{Romdhani:2005} to get the estimated
3D shape information. Then we map those vertices to the $UV$ space
via the cylindrical unwrapping method \cite{Booth:2014}. The
non-visible vertices are culled via z-buffering. As highlighted
above, the training data are not perfectly paired due to the
difficulties in data acquisition, and our goal is to produce
transferred VIS face images with normalized pose. To this end, we
calculate the mean VIS faces for each identity and render the ground
truth $UV$ fields on these mean faces. We find it is a simple yet
effective manner for guiding our approach to estimate the normalized
shape information. Concretely, our $UV$ loss item takes the
following form:

\begin{equation}
{\mathcal{L}_{uv}} = {\left\| {{\mathop{\rm}\nolimits} G_p(\bm{X}) - \overline{\bm{UV}}} \right\|_1},
\end{equation}
where $\overline{\bm{UV}}$ denotes the mean $UV$ field and
\({\left\| \cdot \right\|_1}\) denotes calculating the mean of the
element-wise absolute value summation of a matrix. The $\ell_1$ norm
\({\left\| \cdot \right\|_1}\) can be treated as a robust estimator
to ensure that the majority parts of $G_p(\bm{X})$ and
$\overline{\bm{UV}}$ are close and similar.

\subsection{Transformative Adversarial Texture Inpainting \label{sec:inpa}}
The texture inpainting network $G_t$ aims to encode a given face
texture into a compact identity representation, and then to decode
the representation into the facial texture map in the VIS domain.
Note that in our experiments, the input faces are sampled from both
the NIR and VIS domains. On the one hand, this modification augments
the training data for $G_p$ to learn the pose correction; on the
other hand, we find that making our $G_t$ be aware of the spectrum
of the input also slightly improves the performance. For the input
faces in the VIS domain, $G_t$ will learn the identity mapping. In
order to disentangle the texture representation, $G_t$ should learn
to filter out the other irrelevant information. We accomplish this
goal by introducing $D_t$ as the rival to supervise $G_t$ during the
training process. Concretely, $D_t$ takes a couple of
representations that have the same identity label and predicts
whether the input is a real representation pair. In our experiment,
the representations of the real pair are all from VIS training data.
The fake pair consists of one representation from our synthesized
faces and another one from the VIS training data. $G_t$ tries to
deceive $D_t$ into believing that the fake pair is real. In this
adversarial training scheme, the improvements of $G_t$ are two-fold:
1) $G_t$ will learn to make synthesized results as similar to the
real data in feature space as possible. 2) $G_t$ can eliminate the
intra-class variations of the real data in pose, expression, etc.,
ensuring that the learned representations are closely related to
identity. Formally, the loss item introduced by the adversarial
learning scheme above is as follows:

\begin{equation}
\label{G_t}
\mathcal{L}_{G_{t}}=\mathbb{E}_{\bm{X}\sim p_{data}}[-\log(D_{t}(G_{t}(\bm{X}))].
\end{equation}

In the meantime, the loss item for our discriminator is defined as:

\begin{equation}
\label{D_t}
\mathcal{L}_{D_{t}} = \mathbb{E}_{\bm{X}\sim p_{data}}[-\log(1-D_{t}(G_{t}(\bm{X})) - \log(D_{t}(\bm{X}))].
\end{equation}

\subsection{Fusion Warping Net \label{sec:fuse}}
A fusion warping network is designed to combine the output of $G_p$
and $G_t$, producing final transformed VIS results. It is inspired
by the classical warping operation applied in face manipulation.
Recall that once the $UV$ field is specified, the facial texture map
can be warped into the corresponding 2D face image. However, the
values of pixels standing for background parts, e.g., non-facial
areas, ears and hair, are undefined in this case. Hence, the post
process is necessary to complete the missing parts. To produce the
background parts simultaneously with synthesizing the facial one, we
build the fusion warping net with several convolution layers. The
final output of $D_t$, i.e., the predicted facial texture map,
remains to be warped into the facial region. In the meantime, the
output of the second last layer of $G_t$, which can be regarded as
the facial texture feature map, is fed into the fusion warping net
along with the warped facial part. Besides, the predicted facial
texture map is not limited to being in the RGB color space for our
fusion warping net. In our experiment, we increase the number of
feature channel of the facial texture map to 32 and find that better
performance is obtained.

We introduce the adversarial learning in RGB color space to
supervise the fusion warping net on producing realistic transferred
faces. To achieve high-resolution NIR-VIS face completion, we employ
a multi-scale discriminator, $D_r=\{D_{rl}, D_{rh}\}$. Specifically,
we apply wavelet decomposition on the full-size input data by a
factor of 2, yielding a series of wavelet coefficients. We choose
Haar wavelet because it is enough to depict different-frequency
facial information. We use 2-D fast wavelet transform (FWT)
\cite{Mallat:1989} to compute Haar wavelets. $D_{rl}$ and  $D_{rh}$
aim at discriminating the difference between the real data and the
synthetic results in the low-frequency and the high-frequency
coefficients, respectively. Compared with the single-scale
discriminator, our $D_r$ effectively supervises the generator to
produce both globally and locally consistent results. We find that
when dealing with high-resolution (larger than $256\times256$), the
single-scale discriminator has limited power in generating plausible
local textures. To address this problem, we assign a larger weight
for minimizing the high-frequency adversarial loss. Formally, the
multi-scale adversarial loss for our generator is formulated as:

\begin{equation}
\label{G_F}
\begin{split}
\mathcal{L}_{G_{F}}&=\mathbb{E}_{\bm{X}\sim p_{data}}[-\log(D_{rl}(\phi_{rl}(F(\bm{X}))) \\
&-\lambda\log(D_{rh}(\phi_{rh}(F(\bm{X})))],
\end{split}
\end{equation}
where we denote the output of our fusion warping net as $F(\bm{X})$.
$\phi_{rh}(\cdot)$ and $\phi_{rl}(\cdot)$ denote the decomposed
high-frequency and low-frequency wavelet coefficients, respectively.
We set $\lambda=10$ in our experiment to emphasize generating
plausible high-frequency information.

Correspondingly, the multi-scale adversarial loss for our
discriminator takes the form:
\begin{equation}
\begin{split}
\label{D_F}
\mathcal{L}_{D_{F}} &= \sum_{r}^{\{rl,rh\}}\{\mathbb{E}_{\bm{X}\sim p_{data}}[-\log(D_{r}(\phi_{rl}(\bm{X})))] \\
&-\log(1-D_{r}(G_{r}(\phi_{r}(\bm{X})))\}.
\end{split}
\end{equation}

\subsection{Regularization Items and the Overall Loss Function \label{sec:all}}
Most identity-preserving face generation tasks resort to a
pre-trained face recognizer, e.g., VGG-Face \cite{Parkhi:2015} and
Light CNN \cite{XWuT:2018}, to provide guidance on keeping the
identity information. Specifically, the goal of identity-preserving
is achieved by minimizing the distance between the extracted
identity representations of the real and the synthetic face pair.
Some methods also optimize the distance between the median outputs
provided by the face recognizer, e.g., the output of the last
pooling layer. This regularization item is referred to as the
perceptual loss. We also include this item to improve the
identity-preserving ability of our network, which is formulated as:
\begin{equation}
\label{L_p}
\mathcal{L}_{p}=||\delta(\bm{X})-\delta(F(\bm{X}))||_{2}^{2},
\end{equation}
where $\delta(\cdot)$ denotes the extracted identity representation
obtained by the second last fully connected layer within the
identity preserving network and $||\cdot||_{2}$ means the vector
$\ell_2$-norm.

The pixel-wise losses like $L1$ and $L2$ losses have been proved
effective in keeping low-frequency information. Since the training
set of our NIR-VIS face completion does not have perfectly matched
data pair, heavy reliance on the pixel-wise loss will lead to
over-smooth results. When the resolution increases, this limitation
will be exaggerated and pose significant challenges of producing
plausible texture information. Considering that, we assign a
relatively small factor for pixel-wise loss to help the learning of
global structure. Our pixel-wise loss item takes the following form:
\begin{equation}
\label{L_l1}
\mathcal{L}_{l1}=\alpha||\bm{X}-F(\bm{X})||_{1},
\end{equation}
where we assign $\alpha=0.01$ in our experiment since heavy reliance on the pixel-wise loss significantly harms the visual quality, especially for the high-resolution face completion.

In summary, the overall loss functions for our generator are given
as:
\begin{equation}
\label{L_G}
\mathcal{L}_{G}=\mathcal{L}_{uv}+\mathcal{L}_{G_{t}}\mathcal{+L}_{G_{F}}+\mathcal{L}_{p}+\mathcal{L}_{l1}
\end{equation}
The loss functions for the multi-scale discriminator $D_r$ and the
fine-grained discriminator $D_t$ are $L_{D_t}$ and $L_{D_F}$,
respectively. The generator and the discriminators are iteratively
optimized as suggested by \cite{Goodfellow:2014}. We use the
perceptual loss as the indicator and stop the training process when
it converges.

\begin{figure*}[t]
\begin{center}
    \subfigure[The CASIA NIR-VIS 2.0 database]{\includegraphics[width=0.32\textwidth]{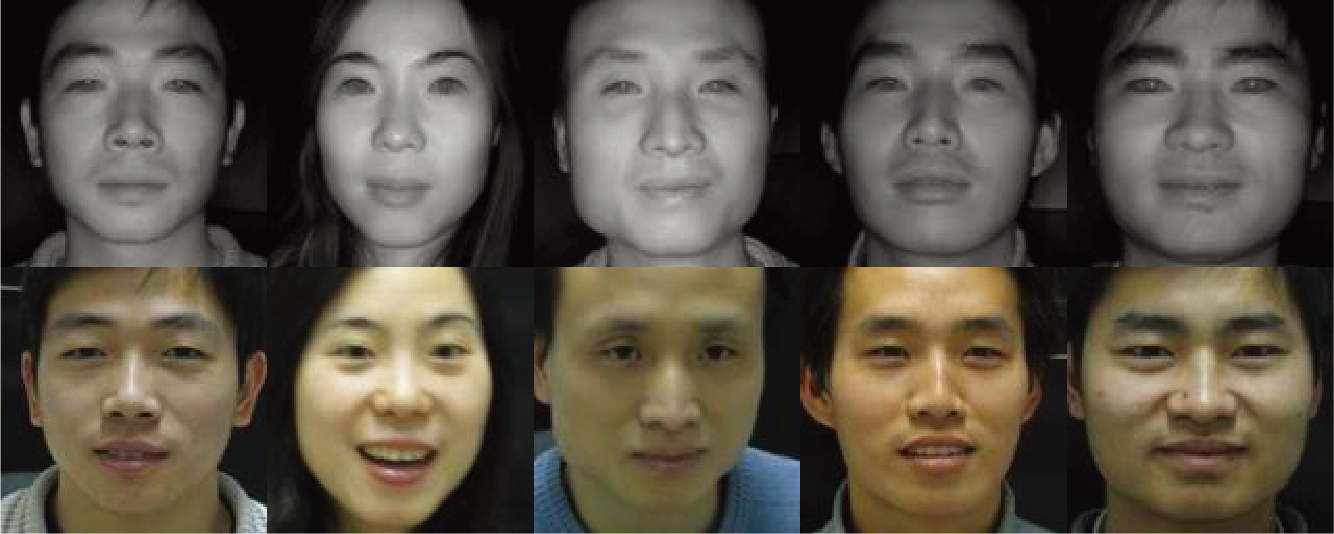}}
    \subfigure[The BUAA-VisNir database]{\includegraphics[width=0.32\textwidth]{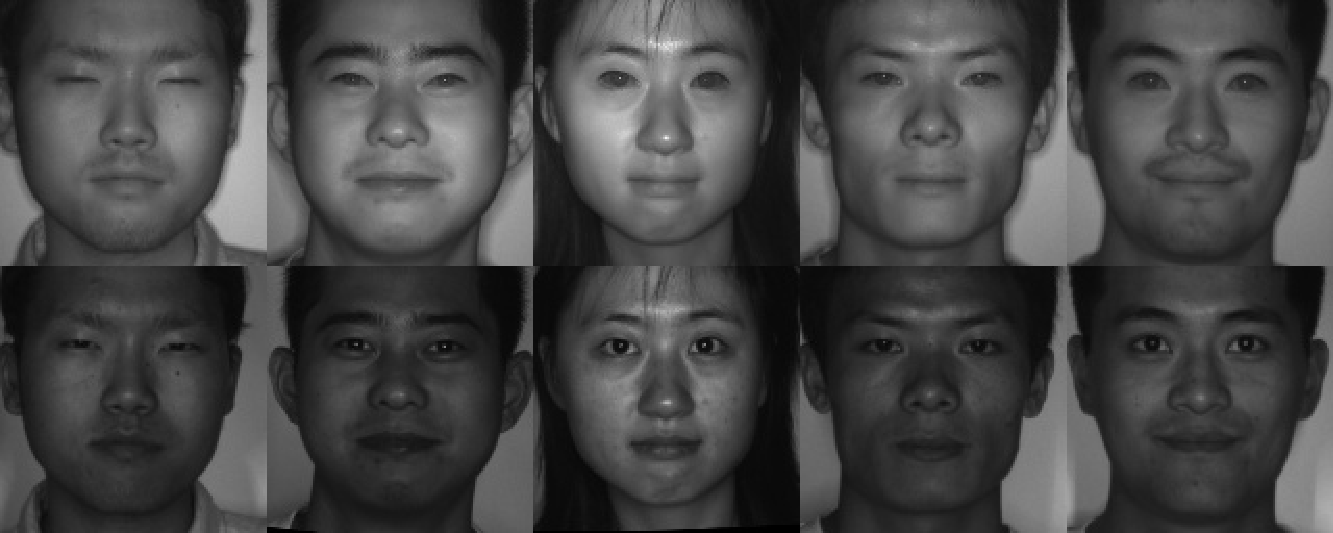}}
    \subfigure[The Oulu-CASIA NIR-VIS database]{\includegraphics[width=0.32\textwidth]{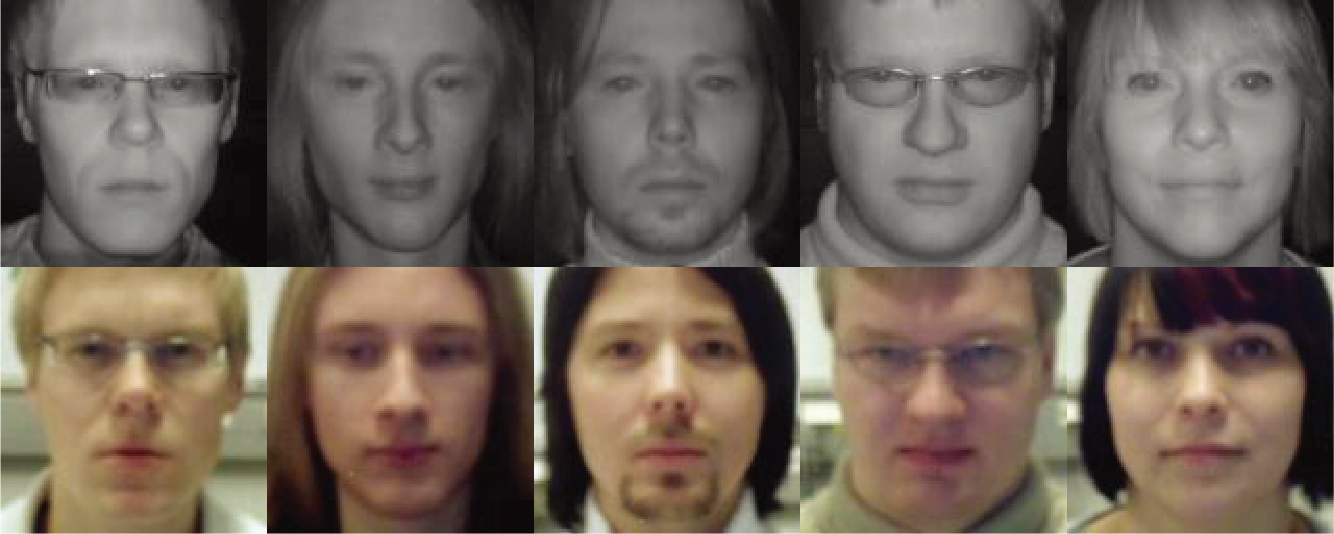}}
\end{center}
   \caption{An illustration of used heterogeneous face images in the three databases. The first row and second row contain a probe NIR image and a VIS gallery image respectively. \label{fig:face}}
\end{figure*}

\subsection{Implementation Details}
Our end-to-end network is implemented based on the deep learning
library Pytorch. An NVIDIA Titan XP GPU with 12GB GDDR5X RAM is
employed for the training and testing processes. We build $G_p$ with
a U-Net structure \cite{Ronneberger:2015} network and adopt the
network structures in \cite{JZhu:2017} to build our $G_t$ and
discriminators. We optimize the parameters of our model by Adam
optimizer \cite{Kingma:2014} with a learning rate of 2e-4 and
momentum of 0.5. We use the perceptual loss as the indicator and
stop training when it no longer decreases. The training processes of
our models producing $128\times128$ and $256\times256$ outputs last
for 3 and 11 hours, respectively.

We employ a pre-trained Light CNN \cite{XWuT:2018} as the baseline
recognizer\footnote{available at
\url{https://github.com/AlfredXiangWu/LightCNN}}. We also use it to
calculate the perceptual loss during the training process. The
network is composed of 29 convolution layers with a variation of
maxout operations, i.e., we use LightCNN-29v2. We train the Light
CNN on MS-Celeb-1M \cite{YDGuo:2106}, which consists of 10K
identities with 8.5M VIS face images. Note that faces in the NIR
domain do not appear in the training process of Light CNN. The
performance of face verification is evaluated by 'recognition via
generation': NIR faces are first processed by our NIR-VIS face
completion method and then fed to the Light CNN for matching. We
also evaluate the verification performance by directly using the NIR
faces as a reference.

\section{Experiments}
In this section, the proposed cross-spectral face completion
framework is systemically evaluated against state-of-the-art HFR
methods and deep learning methods on three widely used HFR face
databases. New benchmark protocols for evaluating 'recognition via
generation' are proposed. Both quantitative and qualitative results
are reported.
\subsection{Experimental Settings}
\subsubsection{Databases}
{\bf The CASIA NIR-VIS 2.0 Face Database}~\cite{SLi:2013} is the
mostly used HFR database (or cross-modal database~\cite{YJin:2017})
because it is the largest public and most challenging HFR database.
It is collected in four recording sessions from 2007 to 2010. There
are large variations of the same identity, including lighting,
expression, pose, and distance. Moreover, wearing glasses or not is
also considered to increase variations. The age distribution of the
subjects spans from children to old people. The total number of the
subjects in this database is 725. Each subject has 1-22 VIS and 5-50
NIR images. Since each image is randomly gathered, NIR and VIS
images have no one-to-one correlations (i.e., they are unpaired).
Fig.~\ref{fig:face} shows some samples of cropped VIS and NIR faces.
We observe that NIR-VIS images are unpaired and have large pose
variations, which make heterogeneous synthesis and matching on this
database challenging. Two views of matching protocols have been used
in this database. View 1 is designed to adjust super-parameter, and
View 2 can be adopted for training and testing.

The matching protocol in View 2 contains 10-fold experiments. In
each fold, there is a collection of training and testing lists. The
training and testing sets in each fold include nearly equal numbers
of identities that are kept disjoint from each other. This means
that the identities used for training and testing are entirely
different, which facilitates a fair comparison of heterogeneous
synthesis and recognition. For each fold, there are about 6,100 NIR
images and 2,500 VIS images from about 360 identities. These
subjects are exclusive from the 358 identities in the testing set.
For the testing of each fold, the gallery contains 358 identities
and each identity has one VIS image. The probe has over 6,000 NIR
images from the same 358 identities. Each NIR image in the probe set
compares against all images in the gallery set, resulting in a
similarity matrix of size $358$ by around $6,000$.


{\bf The BUAA-VisNir face database}~\cite{DHuang:2012} is a standard
HFR database and can also be used as a testing platform for domain
adaptation~\cite{MShao:2014}. It has 150 subjects with 9 VIS images
and 9 NIR images, which are captured simultaneously by using a
single multi-spectral camera. It is often used to evaluate domain
adaptation methods across imaging sensors. The nine images of each
subject are captured under nine distinct poses or expressions, i.e.,
neutral-frontal, left-rotation, right-rotation, tilt-up, tilt-down,
happiness, anger, sorrow and surprise. The training set is composed
of 900 images of 50 subjects and the testing set contains 1800
images from the remaining 100 subjects. To avoid that the probe and
gallery images are in the same pose and expression, we select only
one VIS image of each subject in the gallery set during testing. As
a result, there are 100 VIS images and 900 NIR images in the gallery
set and the probe set respectively. Since there are pose and
illumination variations in the probe set, this testing protocol is
still challenging. Fig.~\ref{fig:face} (b) lists some samples of
cropped VIS and NIR faces.

{\bf The Oulu-CASIA NIR-VIS database}~\cite{JChen:2009} is often
used to study the effects of illumination variations to facial
expressions and heterogeneous face recognition. It contains NIR and
VIS images from 80 subjects. 50 subjects are from Oulu University
and the other 30 subjects are from CASIA. For each subject, there
are six expression variations (anger, disgust, fear, happiness,
sadness, and surprise). NIR and VIS images are acquired under three
different illumination environments: normal indoor, weak and dark.
Fig.~\ref{fig:face} (c) shows some cropped VIS and NIR faces. The
image resolutions of VIS and NIR images are different. Some NIR
images are blurred and low-resolution, which makes HFR more
difficult. Forty subjects are selected for
evaluation~\cite{MShao:2016}, including 10 subjects from Oulu
University and 30 subjects from CASIA 2.0. Eight face images are
randomly selected from each expression for each domain. Then, there
are totally 48 NIR images and 48 VIS images for each subject. 20
subjects are used as training and the remaining 20 subjects are used
as testing. During testing, the gallery set contains all VIS images
of the 20 subjects in testing, and the probe set contains all their
corresponding NIR images.

\begin{figure*}[!t]
\centering
\includegraphics[width=0.95\textwidth]{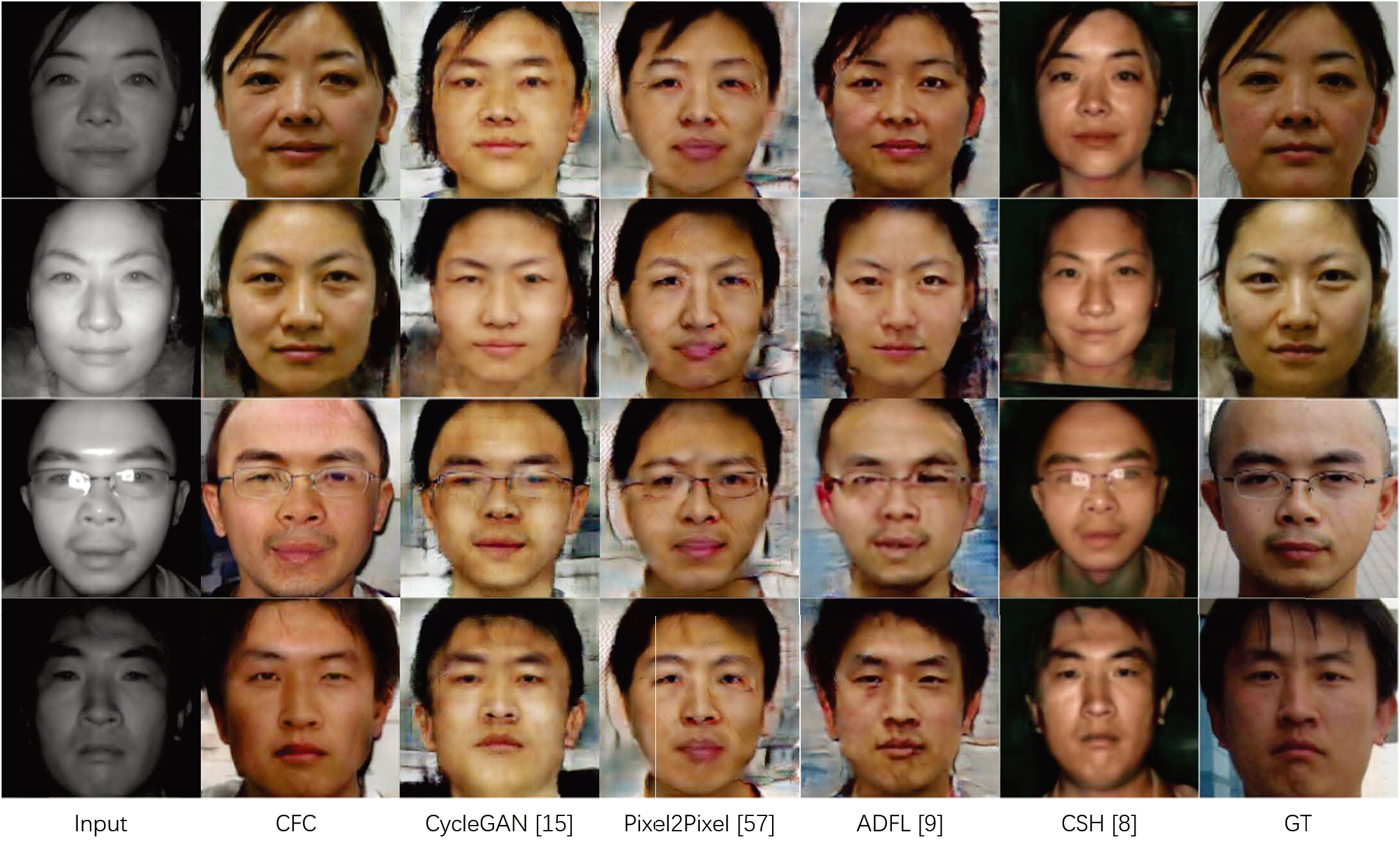}
\caption{Visualization results of different methods. Since the face images in CASIA~2.0 are collected from 2007 to 2010, VIS face images have different skin colors and backgrounds. \label{fig:compare_res}}
\end{figure*}

\subsubsection{Protocols}
Even though there have been some NIR-VIS heterogeneous
databases~\cite{JChen:2009}\cite{DHuang:2012}\cite{SLi:2013}, they
are often used for HFR rather than face synthesis. Since
heterogeneous databases are often small-scale, different synthesis
protocols were adopted to enrich training set
\cite{Lezama:2017}\cite{LXSong:2018}. To the best of our knowledge,
there is still no a benchmark protocol to evaluate 'recognition via
generation' for HFR. To be consistent with the standard 10-fold
protocol in the CASIA NIR-VIS 2.0 database, we define two different
protocols in this paper for further research as follows,

{\bf Synthesis protocol:} Considering that there are only limited
training samples, we employ the NIR and VIS images of all 357
identities in the training set of the first fold of CASIA~2.0 to
train a generative model and the testing protocol of the first fold
to qualitatively evaluate the synthesized results of different
methods\footnote{There are 10 fold experiments on the CASIA 2.0
database. We have trained our model on each fold and only report the
visual results on the first fold.}. There are 6,010 NIR images and
2,547 VIS images from the 357 identities. The identities used for
training and testing are entirely different.

{\bf Recognition protocol:} We follow the 'recognition via
generation' framework~\cite{RHuang:2017} to evaluate recognition
performance. That is, VIS face recognition methods are directly used
to match VIS images and the synthesized images from NIR images. VIS
face recognition methods are not fine-tuned on the synthesized
images. For the CASIA NIR-VIS 2.0 database, we train the generative
model on each fold and use the testing protocol on each fold for
evaluation. The training set is only used to train a generative
model. For the BUAA-VisNir database and the Oulu-CASIA NIR-VIS
database, the training sets in these two databases are not used. We
directly employ the generative model trained on the first fold of
the CASIA NIR-VIS 2.0 database to translate NIR domain to VIS
domain. Then the testing sets in the two databases are used for
evaluation.

\subsection{Face Image Synthesis}
In this subsection, we compare the synthesis results of different
methods. The first fold of CASIA~2.0 is used to train and verify
performance. Pixel2Pixel~\cite{pix2pix2016} and
cycleGAN~\cite{JZhu:2017} are used as baselines. Since
Cross-Spectral Hallucination (CSH)~\cite{Lezama:2017} and
Adversarial Discriminative Feature Learning
(ADFL)~\cite{LXSong:2018} used different training protocols, we only
select the four subjects in the testing set of the first fold of
CASIA~2.0 for visual comparison. These four subjects are also used
in \cite{Lezama:2017}. Hence, the synthesized results of CSH are
directly copied form \cite{Lezama:2017}. Note that the face images
in CASIA~2.0 are collected in four recording sessions (different
environments) that range are from 2007 to 2010~\cite{SLi:2013}.
Hence the VIS face images in CASIA~2.0 have different skin colors
and backgrounds, which make synthesis tasks quite challenging.

Fig.~\ref{fig:compare_res} shows visualization results of different
methods. It is clear that our CFC method significantly outperforms
its competitors. Due to large sensing gap and pose difference,
Pixel2Pixel and cycleGAN can not obtain satisfactory results on the
NIR-VIS heterogeneous problem. Their synthesized results are
consistent with the observation in
\cite{Lezama:2017}\cite{LXSong:2018}. CSH almost loses all
background contents and its synthesized face appearances are quite
similar to those of NIR face images. Since the backgrounds of NIR
images are corrupted and those of VIS images have large variations,
all methods can not perfectly complete background pixels around hair
and ears. Our CFC method seems to generate background pixels better.
We also observe that the synthesized faces by our CFC method have
different skin colors. This is because the face images in CASIA~2.0
are collected in four recording sessions. The ground-truth (denoted
by 'GT') VIS images and input NIR images may be not from the same
recording session.

\begin{table}[htbp]
  \centering
  \caption{The comparison of Rank-1 accuracy (\%) and verification rate (\%) on the CASIA NIR-VIS 2.0 database (the first fold).  \label{tab:gen}}
  \scalebox{0.9}
  {
    \begin{tabular}{cccc}
    \toprule
    Method & Rank-1 & \multicolumn{1}{c}{VR@FAR=1\%} & \multicolumn{1}{c}{VR@FAR=0.1\%} \\
    \midrule
    Pixel2Pixel & 22.13 & 39.22 & 14.45 \\
    CycleGAN & 87.23 & \multicolumn{1}{c}{93.92} & \multicolumn{1}{c}{79.41} \\
    \midrule
    Light CNN & 96.84 & \multicolumn{1}{c}{99.10} & \multicolumn{1}{c}{94.68} \\
    CFC  &  \bf{99.21} & \multicolumn{1}{c}{\bf{99.82}} & \multicolumn{1}{c}{\bf{98.81 }} \\
    \bottomrule
    \end{tabular}%
  }
\end{table}%

\begin{figure}[t]
\begin{center}
    \subfigure[Various poses]{\includegraphics[width=0.48\textwidth]{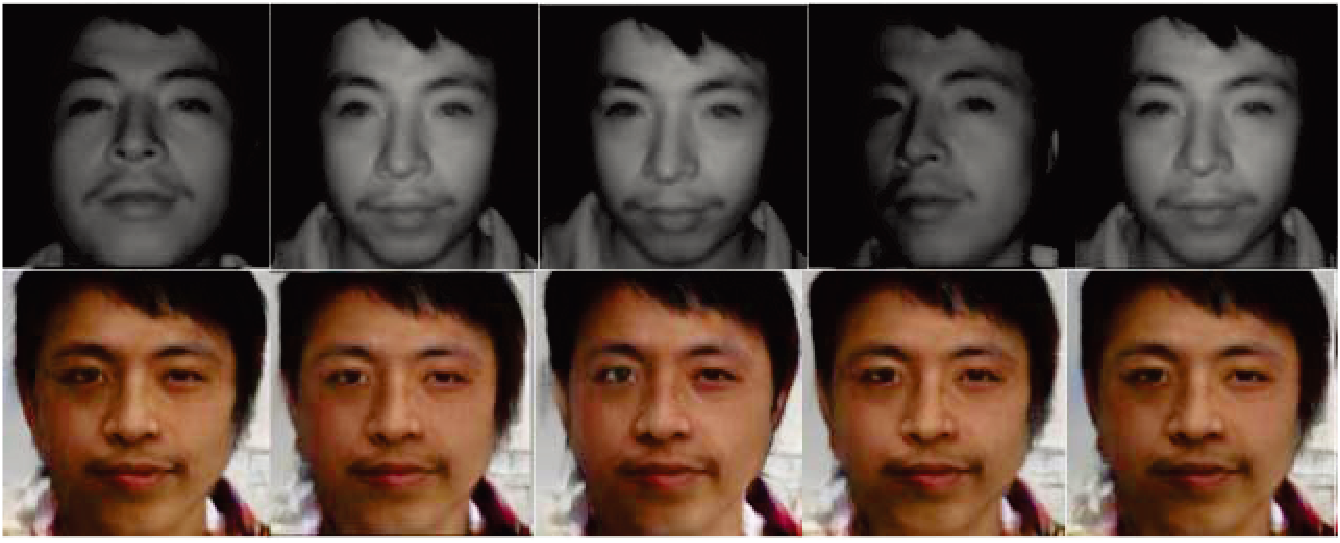}}
    \subfigure[Various expressions]{\includegraphics[width=0.48\textwidth]{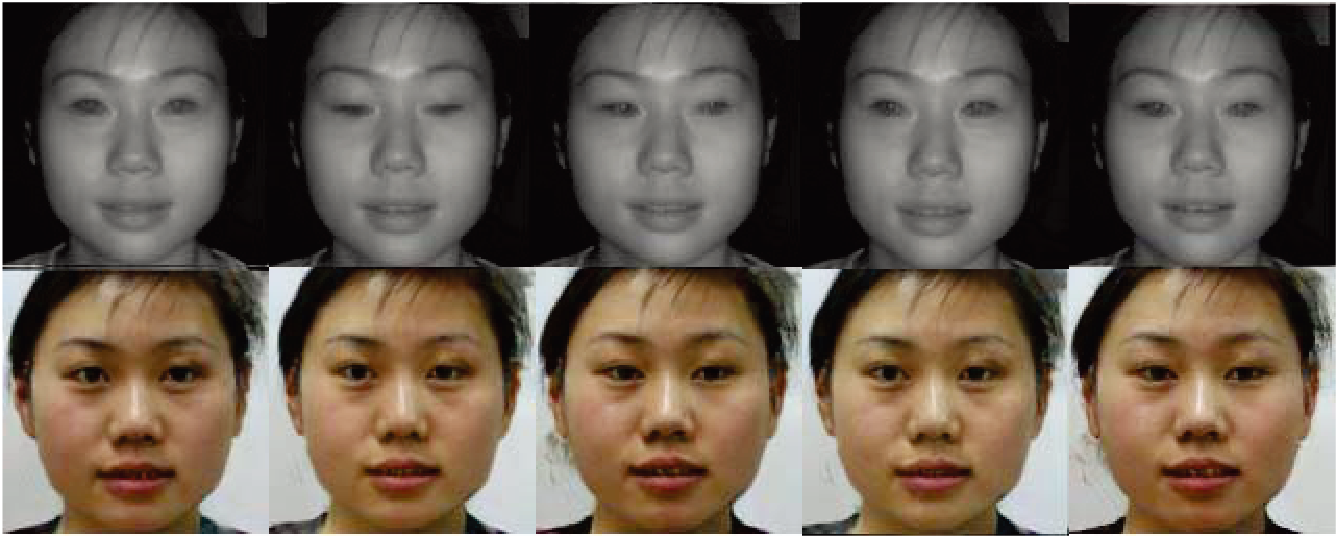}}
\end{center}
   \caption{Synthesized VIS faces (the second row) under different expressions and poses. Our CFC method translates different NIR faces to a frontal VIS face. \label{fig:subject}}
\end{figure}

Table \ref{tab:gen} further lists the quantitative comparison
results of different synthesis methods. As expected, our CFC method
can further improve the matching performance of Light CNN. The
performance of pixel CNN is quite low. This is because NIR-VIS
heterogeneous face synthesis is an unpaired image translation
problem. There is no exact pixel-level correspondence between NIR
and VIS images. We also observe that cycleGAN can not improve the
performance of Light CNN. Although cycleGAN is developed for
unpaired or unsupervised image synthesis, large face variations
(pose, expression) make cycleGAN fail to capture all differences
between NIR and VIS domains. As a result, the HFR performance of
cycle-GAN is not competitive against CFC.

Fig.~\ref{fig:subject} further shows the synthesized VIS face images
from two subjects. There are pose and expression variations in NIR
images, which make face image synthesis and completion challenging.
Our CFC method translates an input NIR face to a frontal VIS face.
Although there are large pose variations, our method can still
generate stable results. There are only large variations on the
generated areas around ear and hair. This may be due to the fact
that the visual background information of a NIR image is often
corrupted. However, the generated VIS face areas are almost similar.
Since CFC can simultaneously reduce sensing gap and pose difference,
its generated VIS images potentially facilitate heterogeneous
matching.

\begin{table}[htbp]
  \centering
  \caption{The comparison of Rank-1 accuracy (\%) and verification rate (\%) on the CASIA NIR-VIS 2.0 database. (10-fold) \label{tab:CASIA}}
    \begin{tabular}{cccc}
    \toprule
    Method & Rank-1  & \multicolumn{1}{c}{VR@FAR=1\%} & \multicolumn{1}{c}{VR@FAR=0.1\%} \\
    \midrule
    DSIFT   & 73.3$\pm$1.10 & \multicolumn{1}{c}{-} & \multicolumn{1}{c}{-} \\
    CDFL   & 71.5$\pm$1.40 & \multicolumn{1}{c}{67.7} & \multicolumn{1}{c}{55.1} \\
    Gabor+RBM   & 86.2$\pm$0.98 & \multicolumn{1}{c}{-} & \multicolumn{1}{c}{81.3$\pm$1.82} \\
    LCFS   & 35.4$\pm$2.80 & \multicolumn{1}{c}{35.7} & \multicolumn{1}{c}{16.7} \\
    H2(LBP3)   & 43.8 & \multicolumn{1}{c}{36.5} & \multicolumn{1}{c}{10.1} \\
    CEFD   & 85.6 & \multicolumn{1}{c}{-} & \multicolumn{1}{c}{-} \\
    HFR-CNN   & 85.9$\pm$0.90 & \multicolumn{1}{c}{-} & \multicolumn{1}{c}{78.0} \\
    TRIVET  & 95.7$\pm$0.52 & \multicolumn{1}{c}{98.1$\pm$0.31} & \multicolumn{1}{c}{91.0$\pm$1.26} \\
    IDNet   & 87.1$\pm$0.88 & \multicolumn{1}{c}{-} & \multicolumn{1}{c}{74.5} \\
    IDR-128 & 97.3$\pm$0.43 & \multicolumn{1}{c}{98.9$\pm$0.29} & \multicolumn{1}{c}{95.7$\pm$0.73} \\
    ADFL    & 98.2$\pm$0.34 & 99.1$\pm$0.15 & 97.2$\pm$0.48\\
    \midrule
    VGG   & 62.1$\pm$1.88 & \multicolumn{1}{c}{71.0$\pm$1.25} & \multicolumn{1}{c}{39.7$\pm$2.85} \\
    SeetaFace & 68.0$\pm$1.66 & \multicolumn{1}{c}{85.2$\pm$1.13} & \multicolumn{1}{c}{58.8$\pm$2.26} \\
    CenterLoss & 87.7$\pm$1.45 & \multicolumn{1}{c}{88.7$\pm$1.21} & \multicolumn{1}{c}{69.7$\pm$2.07} \\
    Light CNN & 96.7$\pm$0.23 & \multicolumn{1}{c}{98.5$\pm$0.64} & \multicolumn{1}{c}{94.8$\pm$0.43} \\
    \midrule
    CFC  &  98.6$\pm$0.12 & \multicolumn{1}{c}{99.2$\pm$0.08} & \multicolumn{1}{c}{97.3$\pm$0.17} \\
    CFC-Fuse  &  \bf{99.5$\pm$0.10} & \bf{99.8$\pm$0.09} & \bf{97.5$\pm$0.19} \\
    \bottomrule
    \end{tabular}%
\end{table}%

\subsection{NIR-VIS Face Recognition}
Table~\ref{tab:CASIA} lists recognition results on the CASIA NIR-VIS
2.0 Database. Recently proposed state-of-the-art HFR methods are
compared, including six traditional methods and nine deep learning
methods. The traditional methods include Learning Coupled Feature
Spaces (LCFS) \cite{KWang:2016}, DSIFT \cite{Dhamecha:2014}, Coupled
Discriminant Feature Learning (CDFL) \cite{YJin:2015},
Gabor+RBM\cite{DYi:2015}, H2(LBP3) \cite{MShao:2016}, common
encoding feature discriminant (CEFD)~\cite{DGong:2017}. The results
of LCFS and CDFL are from \cite{YJin:2015}, and those of the
remaining compared traditional methods are from their published
papers. For deep learning methods, we compare the recently proposed
TRIVET\cite{XXLiu:2016}, HFR-CNNs \cite{Saxena:2016}, IDNet
\cite{Reale:2016}, Invariant Deep Representation (IDR)
\cite{RHe:2017} and Adversarial Discriminative Feature Learning
(ADFL) \cite{LXSong:2018}. Moreover, the results of three VIS CNN
methods are also discussed, including VGG \cite{Parkhi:2015},
SeetaFace \cite{XLiu:2016} and CenterLoss \cite{YWen:2016}. The
result of Light CNN is used as the baseline of deep methods. Since
most of HFR methods use the standard protocol in View 2 for
evaluation, the results of compared methods are directly reported
from their published papers.

We observe that some deep learning methods significantly outperform
traditional HFR methods. However, some deep learning methods
(including VGG, SeetaFace, and HFR-CNNs) even perform worse than the
traditional Gabor+RBM method in terms of rank-1 accuracy. This may
be because the sensing gap and the over-fitting problem on
small-scale datasets make HFR challenging for deep learning methods.
Different from other methods, our cross-spectral face completion
method does not fine-tune a VIS CNN model on heterogeneous datasets.
CFC directly employs Light CNN to match VIS faces against
synthesized images. It further improves the rank-1 accuracy of Light
CNN from $96.7\%$ to $98.6\%$. Compared to other deep learning, CFC
also achieves comparable recognition performance in terms of Rank-1
accuracy and verification rates. The performances of CFC and ADFL
are close. However, ADFL needs to fine VIS recognition models on
NIR-VIS datasets to improve accuracy. CFC-Fuse further fuses the
features of the original NIR image and its synthesized VIS image.
Both the two features are extracted by Light CNN. We use the mean
value of two features as the feature of CFC-Fuse. It is obvious that
the fusion strategy can further improve performance. These results
suggest that the sensing gap between NIR and VIS domains can be
bridged in a synthesis way.

\begin{table}[htbp]
  \centering
  \caption{Rank-1 accuracy and verification rate on the BUAA NIR-VIS Database. \label{tab:buaa}}
    \begin{tabular}{cccc}
    \toprule
    Method & Rank-1 & FAR=1\% & FAR=0.1\% \\
    \midrule
    MPL3  & 53.2  & 58.1  & 33.3 \\
    KCSR  & 81.4  & 83.8  & 66.7 \\
    KPS   & 66.6  & 60.2  & 41.7 \\
    KDSR  & 83    & 86.8  & 69.5 \\
    H2(LBP3) & 88.8  & 88.8  & 73.4 \\
    TRIVET & 93.9  & 93.0    & 80.9 \\
    IDR   & 94.3  & 93.4  & 84.7 \\
    ADFL  & 95.2  &95.3   &88.0\\
    Light CNN  & 96.5 & 95.4 & 86.7 \\
    \midrule
    CFC  & \bf{99.7}  & \bf{98.7}  & \bf{97.8} \\
    \bottomrule
    \end{tabular}%
\end{table}%

The proposed method is further evaluated on the BUAA VisNir
database. The recognition testing protocol in \cite{MShao:2016} is
used. We compare CFC with MPL3 \cite{JChen:2009}, KCSR
\cite{ZLei:2009}, KPS \cite{Klare:2013}, KDSR \cite{XHuang:2013},
KDSR \cite{XHuang:2013} and H2(LBP3) \cite{MShao:2016}. The results
of MPL3, KCSR, KPS, KDSR and H2(LBP3) are from \cite{MShao:2016}.
TRIVET, IDR and ADFL are used as the baseline of deep learning
methods. Table~\ref{tab:buaa} shows Rank-1 accuracy and verification
rates of different HFR methods. We observe that the deep learning
methods outperform the traditional methods in terms of both rank
accuracy and verification rates. Our CFC significantly outperforms
its competitors. These performance improvements of CFC partly
benefit from the usage of Light CNN. Light CNN is trained on a
large-scale VIS dataset so that it can better capture intra-class
variations to facilitate face recognition. Compared to Light CNN, we
can observe that CFC improves the rank-1 accuracy from 96.5\% to
99.7\% and VR@FAR=0.1 from 86.7\% to 97.8\%. These improvements
further suggest that CFC can reduce the sensing gap.

\begin{figure*}[!t]
\centering
\includegraphics[width=0.95\textwidth]{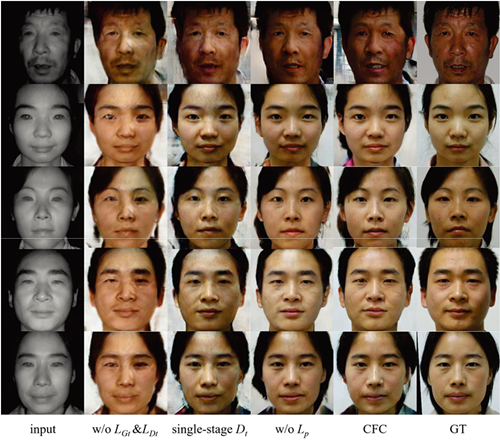}
\caption{Visualization results of high-resolution ($256 \times 256$) NIR-VIS face completion under different configurations. Since background pixels are corrupted in NIR images, there are large variations on the synthesized background areas by different configurations. \label{fig:visual_res}}
\end{figure*}

Table~\ref{tab:oulu} lists HFR performance of different methods on
the Oulu-CASIA NIR-VIS Database. The results of MPL3, KCSR, KPS,
KDSR and H2(LBP3) are from \cite{MShao:2016}. It is observed that
the HFR methods can be ordered in ascending rank-1 accuracy as MPL3,
KPS, KCSR, KDSR, H2(LBP3), TRIVET, IDR, ADFL and CFC. As expected,
CFC achieves the highest rank-1 accuracy (100\%). That is each probe
NIR image is correctly matched to its corresponding VIS image from
the same identity. To the best of our knowledge, it is the first
time that such high rank-1 accuracy is achieved by using a image
synthesis method. Although CFC achieves high rank-1 accuracy on this
database, the verification rates of CFC on the Oulu-CASIA NIR-VIS
database are lower than those of CFC on the previous databases. The
lower accuracy of CFC can be attributed to the factor that the NIR
images in this database are blurred and low-resolution, resulting in
large and unpredictable domain difference. This factor also makes
TRIVET and IDR only slightly outperform five traditional methods at
a low FAR.

\begin{table}[htbp]
  \centering
  \caption{Rank-1 accuracy and verification rate on the Oulu-CASIA NIR-VIS Database. \label{tab:oulu}}
    \begin{tabular}{cccc}
    \toprule
    Method & Rank-1 & FAR=1\% & FAR=0.1\% \\
    \midrule
    MPL3  & 48.9  & 41.9  & 11.4 \\
    KCSR  & 66    & 49.7  & 26.1 \\
    KPS   & 62.2  & 48.3  & 22.2 \\
    KDSR  & 66.9  & 56.1  & 31.9 \\
    H2(LBP3) & 70.8  & 62.0    & 33.6 \\
    TRIVET & 92.2 & 67.9  & 33.6 \\
    IDR   & 94.3  & 73.4  & 46.2 \\
    ADFL  & 95.5 & 83.0   & 60.7 \\
    Light CNN & 96.7 & 92.4 & 65.1 \\
    \midrule
    CFC  & \bf{99.9} & \bf{98.1} & \bf{90.7} \\
    \bottomrule
    \end{tabular}%
\end{table}%

\subsection{Ablation Study}
In order to demonstrate the compelling perceptual results generated
by our CFC method as well as the contribution of each component, we
visualize high-resolution NIR-VIS results under different
configurations. Beyond the light spectrum difference, there are also
pose, expression, age and shape variations between NIR and VIS
images. Fig.~\ref{fig:visual_res} shows visualization results of
high-resolution ($256 \times 256$) NIR-VIS face generation under
different configurations. The first column and last column contain
input NIR images and ground-truth VIS images respectively. Note that
the ground-truth image may be not captured under the same time with
the NIR input image. It is interesting to find that there are large
variations on the generated background areas by different
configurations. This may be because background pixels are often
corrupted in NIR images and some information (e.g., hair texture) is
missed.

We observe that all components of our CFC method contribute to a
high-resolution and high-quality result. The notation 'w/o
$L_{G_F},L_{D_F}$' indicates that CFC only uses the pixel loss and
does not contain a discriminator on the final synthesized RGB face
image. It is clear that the images in the second column are more
blur than those in the fifth column. Although CFC has also a
discriminator during adversarial texture inpainting, the
discriminator w.r.t. $L_{G_F}, L_{D_F}$ plays an important role
during face synthesis. 'single-scale $D_t$' indicates that we
replace the multi-scale $D_t$ of CFC with a single-scale one. That
is, we do not perform a discriminator on Wavelet decomposition. It
seems that the images in the third column have fewer local details
than the images in the fifth column. These results demonstrate the
effectiveness of wavelet based multi-scale discriminators. The
notation 'w/o $L_p$' indicates the CFC model without using identity
preserving loss $L_p$. Comparing the images in the fourth column and
fifth column, we observe that there are obvious differences between
the pixels around ear and hair. It seems that when the identity
preserving loss is used, the pixels around ear and hair are better
generated. Since the background pixels are corrupted in NIR images,
identity preserving loss potentially ensures that CFC can better
complete missing information during synthesis.

\section{Conclusion}
In this paper, we have proposed a new architecture for NIR-VIS face
image synthesis. Different from previous synthesis methods for HFR,
we model the heterogeneous synthesis using two complementary
components: a texture inpainting component and a pose correction
component. The synthesis problem is simplified into two learning
problems, facilitating one-to-one supervised texture completion.
Then a warping procedure is used to fuse the two components in an
end-to-end deep network. A multi-scale discriminator and a
fine-grained discriminator have been developed to improve image
quality. A new benchmark on CASIA NIR-VIS 2.0 has been established
to systematically evaluate the synthesis results for HFR
performance. Extensive experimental results on cross-database
validation show that our network not only generates high-resolution
face images and but also facilitates the accuracy improvements of
state-of-the-art heterogeneous face recognition.



%

%

\ifCLASSOPTIONcompsoc
  \section*{Acknowledgments}
\else
  \section*{Acknowledgment}
\fi

This work is funded by State Key Development Program (Grant No. 2016YFB1001001), the National Natural Science Foundation of China (Grants No. 61622310), and Beijing Natural Science Foundation (Grants No. JQ18017).

\ifCLASSOPTIONcaptionsoff
  \newpage
\fi



\bibliographystyle{IEEEtran}
{
\bibliography{reference}

\begin{thebibliography}{10}
\providecommand{\url}[1]{#1}
\csname url@rmstyle\endcsname
\providecommand{\newblock}{\relax}
\providecommand{\bibinfo}[2]{#2}
\providecommand\BIBentrySTDinterwordspacing{\spaceskip=0pt\relax}
\providecommand\BIBentryALTinterwordstretchfactor{4}
\providecommand\BIBentryALTinterwordspacing{\spaceskip=\fontdimen2\font plus
\BIBentryALTinterwordstretchfactor\fontdimen3\font minus
  \fontdimen4\font\relax}
\providecommand\BIBforeignlanguage[2]{{%
\expandafter\ifx\csname l@#1\endcsname\relax
\typeout{** WARNING: IEEEtran.bst: No hyphenation pattern has been}%
\typeout{** loaded for the language `#1'. Using the pattern for}%
\typeout{** the default language instead.}%
\else
\language=\csname l@#1\endcsname
\fi
#2}}

\bibitem{SOuyang:2016}
S.~Ouyang, T.~Hospedales, Y.-Z. Song, X.~Li, C.~C. Loy, and X.~Wang, ``A survey
  on heterogeneous face recognition: Sketch, infra-red, 3{D} and
  low-resolution,'' \emph{Image and Vision Computing}, vol.~56, pp. 28--48,
  2016.

\bibitem{RHe:2018}
R.~He, X.~Wu, Z.~Sun, and T.~Tan, ``Wasserstein {CNN}: Learning invariant
  features for {NIR}-{VIS} face recognition,'' \emph{IEEE Transactions on
  Pattern Analysis and Machine Intelligence}, vol.
  DOI:10.1109/TPAMI.2018.2842770, 2018.

\bibitem{Reale:2016}
C.~Reale, N.~M. Nasrabadi, H.~Kwon, and R.~Chellappa, ``Seeing the forest from
  the trees: A holistic approach to near-infrared heterogeneous face
  recognition,'' in \emph{IEEE Workshop on Perception Beyond the Visible
  Spectrum}, 2016, pp. 54--62.

\bibitem{RHe:2017}
R.~He, X.~Wu, Z.~Sun, and T.~Tan, ``Learning invariant deep representation for
  {NIR}-{VIS} face recognition,'' in \emph{AAAI Conference on Artificial
  Intelligence}, 2017, pp. 2000--2006.

\bibitem{Sarfraz:2017}
M.~S. Sarfraz and R.~Stiefelhagen, ``Deep perceptual mapping for cross-modal
  face recognition,'' \emph{International Journal of Computer Vision}, vol.
  122, no.~3, pp. 426--438, 2017.

\bibitem{Riggan:2016}
B.~Riggan, N.~Short, S.~Hu, and H.~Kwon, ``Estimation of visible spectrum faces
  from polarimetric thermal faces,'' in \emph{IEEE International Conference on
  Biometrics Theory, Applications and Systems}, 2016, pp. 1--7.

\bibitem{HZhang:2017}
H.~Zhang, V.~Patel, B.~Riggan, and S.~Hu, ``Generative adversarial
  network-based synthesis of visible faces from polarimetric thermal faces,''
  in \emph{International Joint Conference on Biometrics}, 2017.

\bibitem{Lezama:2017}
J.~Lezama, Q.~Qiu, and G.~Sapiro, ``Not afraid of the dark: {NIR}-{VIS} face
  recognition via cross-spectral hallucination and low-rank embedding,'' in
  \emph{IEEE Conference on Computer Vision and Pattern Recognition}, 2017.

\bibitem{LXSong:2018}
L.~Song, M.~Zhang, XiangWu, and R.~He, ``Adversarial discriminative
  heterogeneous face recognition,'' in \emph{AAAI Conference on Artificial
  Intelligence}, 2018.

\bibitem{Goodfellow:2014}
I.~J. Goodfellow, J.~Pouget-Abadie, M.~Mirza, B.~Xu, D.~Warde-Farley, S.~Ozair,
  A.~Courville, and Y.~Bengio, ``Generative adversarial networks,'' in
  \emph{Neural Information Processing System}, 2014.

\bibitem{JDeng:2018}
J.~Deng, S.~Cheng, N.~Xue, Y.~Zhou, and S.~Zafeiriou, ``{UV}-{GAN}: Adversarial
  facial {UV} map completion for pose-invariant face recognition,'' in
  \emph{IEEE Conference on Computer Vision and Pattern Recognition}, 2018.

\bibitem{SLi:2013}
S.~Z. Li, D.~Yi, Z.~Lei, and S.~Liao, ``The casia {nir}-{vis} 2.0 face
  database,'' in \emph{IEEE Conference on Computer Vision and Pattern
  Recognition Workshops}, 2013, pp. 348--353.

\bibitem{DHuang:2012}
D.~Huang, J.~Sun, and Y.~Wang, ``The {BUAA}-{V}is{N}ir face database
  instructions,'' Beihang University, Beijing, China, Tech. Rep.
  IRIP-TR-12-FR-001, July 2012.

\bibitem{JChen:2009}
J.~Chen, D.~Yi, J.~Yang, G.~Zhao, S.~Z. Li, and M.~Pietikainen, ``Learning
  mappings for face synthesis from near infrared to visual light images,'' in
  \emph{IEEE Conference Computer Vision Pattern Recognition}, 2009, pp.
  156--163.

\bibitem{JZhu:2017}
J.~Zhu, T.~Park, P.~Isola, and A.~Efros, ``Unpaired image-to-image translation
  using cycle-consistent adversarial networks,'' in \emph{IEEE International
  Conference on Computer Vision}, 2017.

\bibitem{SLi:2006}
S.~Z. Li, R.~Chu, M.~Ao, L.~Zhang, and R.~He, ``Highly accurate and fast face
  recognition using near infrared images,'' in \emph{IAPR International
  Conference on Biometrics}, 2006, pp. 151--158.

\bibitem{LXiao:2013}
L.~Xiao, R.~He, Z.~Sun, and T.~Tan, ``Coupled feature selection for
  cross-sensor iris recognition,'' in \emph{IEEE conference on Biometrics:
  Theory, Applications and Systems}, 2013, pp. 1--6.

\bibitem{JZhu:2014}
J.-Y. Zhu, W.-S. Zheng, J.-H. Lai, and S.~Z. Li, ``Matching {NIR} face to {VIS}
  face using transduction,'' \emph{IEEE Transactions on Information Forensics
  and Security}, vol.~9, no.~3, pp. 501--514, 2014.

\bibitem{RWang:2009}
R.~Wang, J.~Yang, D.~Yi, and S.~Li, ``An analysis-by-synthesis method for
  heterogeneous face biometrics,'' in \emph{IAPR International Conference on
  Biometrics}, 2009, pp. 319--326.

\bibitem{XTang:2003}
X.~Tang and X.~Wang, ``Face sketch synthesis and recognition,'' in \emph{IEEE
  Conference on Computer Vision and Pattern Recognition}, 2003, pp. 687--694.

\bibitem{XWang:2009}
X.~Wang and X.~Tang, ``Face photo-sketch synthesis and recognition,''
  \emph{IEEE Transactions on Pattern Analysis and Machine Intelligence},
  vol.~31, no.~11, pp. 1955--1967, 2009.

\bibitem{ZLei:2008r}
Z.~Lei, Q.~Bai, R.~He, and S.~Li, ``Face shape recovery from a single image
  using cca mapping between tensor spaces,'' in \emph{IEEE Conference on
  Computer Vision and Pattern Recognition}, 2008.

\bibitem{SWang:2012}
S.~Wang, D.~Zhang, Y.~Liang, and Q.~Pan, ``Semi-coupled dictionary learning
  with applications to image super-resolution and photosketch synthesis,'' in
  \emph{IEEE Conference on Computer Vision and Pattern Recognition}, 2012, pp.
  2216--2223.

\bibitem{DHuang:2013}
D.-A. Huang and Y.-C.~F. Wang, ``Coupled dictionary and feature space learning
  with applications to cross-domain image synthesis and recognition,'' in
  \emph{IEEE International Conference on Computer Vision}, 2013, pp.
  2496--2503.

\bibitem{FXu:2015}
F.~Juefei-Xu, D.~K. Pal, and M.~Savvides, ``{NIR}-{VIS} heterogeneous face
  recognition via cross-spectral joint dictionary learning and
  reconstruction,'' in \emph{IEEE Conference on Computer Vision and Pattern
  Recognition Workshop}, 2015.

\bibitem{MZhang:2019}
M.~Zhang, R.~Wang, X.~Gao, J.~Li, and D.~Tao, ``Dual-transfer face
  sketch¨cphoto synthesis,'' \emph{IEEE Transactions on Image Processing},
  vol.~28, no.~2, pp. 642--657, 2019.

\bibitem{RHuang:2017}
R.~Huang, S.~Zhang, T.~Li, and R.~He, ``Beyond face rotation: Global and local
  perception gan for photorealistic and identity preserving frontal view
  synthesis,'' in \emph{IEEE International Conference on Computer Vision},
  2017, pp. 2458--2467.

\bibitem{DLin:2006}
D.~Lin and X.~Tang, ``Inter-modality face recognition,'' in \emph{IEEE Europe
  Conference on Computer Vision}, 2006, pp. 13--26.

\bibitem{ZLei:2012}
Z.~Lei, S.~Liao, A.~K. Jain, and S.~Z. Li, ``Coupled discriminant analysis for
  heterogeneous face recognition,'' \emph{IEEE Transactions on Information
  Forensics and Security}, vol.~7, no.~6, pp. 1707--1716, 2012.

\bibitem{XHuang:2013}
X.~Huang, Z.~Lei, M.~Fan, X.~Wang, and S.~Z. Li, ``Regularized discriminative
  spectral regression method for heterogeneous face matching,'' \emph{IEEE
  Transactions on Image Processing}, vol.~22, no.~1, pp. 353--362, 2013.

\bibitem{KWang:2016}
K.~Wang, R.~He, L.~Wang, W.~Wang, and T.~Tan, ``Joint feature selection and
  subspace learning for cross-modal retrieval,'' \emph{IEEE Transactions on
  Pattern Analysis and Machine Intelligence}, vol.~38, no.~10, pp. 2010--2023,
  2016.

\bibitem{Klare:2013}
B.~F. Klare and A.~K. Jain, ``Heterogeneous face recognition using kernel
  prototype similarities,'' \emph{IEEE Transactions on Pattern Analysis and
  Machine Intelligence}, vol.~35, no.~6, pp. 1410--1422, 2013.

\bibitem{CHou:2014}
C.-A. Hou, M.-C. Yang, and Y.-C.~F. Wang, ``Domain adaptive self-taught
  learning for heterogeneous face recognition,'' in \emph{International
  Conference on Pattern Recognition}, 2014.

\bibitem{MShao:2014}
M.~Shao, D.~Kit, and Y.~Fu, ``Generalized transfer subspace learning through
  low-rank constraint,'' \emph{International Journal of Computer Vision}, vol.
  109, no. 1-2, pp. 74--93, 2014.

\bibitem{DYi:2015}
D.~Yi, Z.~Lei, S.~Liao, and S.~Li, ``Shared representation learning for
  heterogeneous face recognition,'' in \emph{IEEE International Conference and
  Workshops on Automatic Face and Gesture Recognition}, 2015.

\bibitem{MKan:2016}
M.~Kan, S.~Shan, H.~Zhang, S.~Lao, and X.~Chen, ``Multi-view discriminant
  analysis,'' \emph{IEEE Transactions on Pattern Analysis and Machine
  Intelligence}, vol.~38, no.~11, pp. 188--194, 2016.

\bibitem{ZFLi:2016}
Z.~Li, D.~Gong, Q.~Li, D.~Tao, and X.~Li, ``Mutual component analysis for
  heterogeneous face recognition,'' \emph{ACM Transactions on Intelligent
  Systems and Technology}, vol.~7, no.~3, pp. 1--23, 2016.

\bibitem{YJin:2017}
Y.~Jin, J.~Li, C.~Lang, and Q.~Ruan, ``Multi-task clustering {ELM} for
  {VIS}-{NIR} cross-modal feature learning,'' \emph{Multidimensional Systems
  and Signal Processing}, vol.~28, no.~3, pp. 905--920, 2017.

\bibitem{JGui:2018}
J.~Gui and P.~Li, ``Multi-view feature selection for heterogeneous face
  recognition,'' in \emph{IEEE International Conference on Data Mining}, 2018,
  pp. 983--988.

\bibitem{SLiao:2009}
S.~Liao, D.~Yi, Z.~Lei, R.~Qin, and S.~Z. Li, ``Heterogeneous face recognition
  from local structures of normalized appearance,'' in \emph{IAPR International
  Conference on Biometrics}, 2009, pp. 209--218.

\bibitem{Klare:2011}
B.~F. Klare, Z.~Li, and A.~K. Jain, ``Matching forensic sketches to mug shot
  photos,'' \emph{IEEE Transactions on Pattern Analysis and Machine
  Intelligence}, vol.~33, no.~3, pp. 639--646, 2011.

\bibitem{Goswami:2011}
D.~Goswami, C.~H. Chan, D.~Windridge, and J.~Kittler, ``Evaluation of face
  recognition system in heterogeneous environments (visible vs {NIR}),'' in
  \emph{IEEE International Conference on Computer Vision Workshop}, 2011, pp.
  2160--2167.

\bibitem{Dhamecha:2014}
T.~I. Dhamecha, P.~Sharma, R.~Singh, and M.~Vatsa, ``On effectiveness of
  histogram of oriented gradient features for visible to near infrared face
  matching,'' in \emph{International Conference on Pattern Recognition}, 2014,
  pp. 1788--1793.

\bibitem{MShao:2016}
M.~Shao and Y.~Fu, ``Cross-modality feature learning through generic
  hierarchical hyperlingual-words,'' \emph{IEEE Transactions on Neural Networks
  and Learning Systems}, vol.~28, no.~2, pp. 451--463, 2016.

\bibitem{DGong:2017}
D.~Gong, Z.~Li, W.~Huang, X.~Li, and D.~Tao, ``Heterogeneous face recognition:
  A common encoding feature discriminant approach,'' \emph{IEEE Transactions on
  Image Processing}, vol.~26, no.~5, pp. 2079--2089, 2017.

\bibitem{Saxena:2016}
S.~Saxena and J.~Verbeek, ``Heterogeneous face recognition with {CNN}s,'' in
  \emph{European Conference on Computer Vision Workshops}, 2016, pp. 483--491.

\bibitem{XXLiu:2016}
X.~Liu, L.~Song, X.~Wu, and T.~Tan, ``Transferring deep representation for
  {NIR}-{VIS} heterogeneous face recognition,'' in \emph{IAPR International
  Conference on Biometrics}, 2016.

\bibitem{XWu:2019}
X.~Wu, H.~Huang, V.~M. Patel, R.~He, and Z.~Sun, ``Disentangled variational
  representation for heterogeneous face recognition,'' in \emph{AAAI Conference
  on Artificial Intelligence}, 2019.

\bibitem{Paysan:2009}
P.~Paysan, R.~Knothe, B.~Amberg, S.~Romdhani, and T.~Vetter, ``A {3D} face
  model for pose and illumination invariant face recognition,'' in \emph{IEEE
  International Conference on Advanced Video and Signal Based Surveillance},
  2009, pp. 296--301.

\bibitem{Romdhani:2005}
S.~Romdhani and T.~Vetter, ``Estimating {3D} shape and texture using pixel
  intensity, edges, specular highlights, texture constraints and a prior,'' in
  \emph{IEEE Conference on Computer Vision and Pattern Recognition}, 2005, pp.
  986--993.

\bibitem{Booth:2014}
J.~Booth and S.~Zafeiriou, ``{Optimal {UV} spaces for facial morphable model
  construction},'' in \emph{IEEE International Conference on Image Processing},
  2014, pp. 4672--4676.

\bibitem{Mallat:1989}
S.~G. Mallat, ``A theory for multiresolution signal decomposition: the wavelet
  representation,'' \emph{IEEE Transactions on Pattern Analysis and Machine
  Intelligence}, vol.~11, no.~7, pp. 674--693, 1989.

\bibitem{Parkhi:2015}
O.~M. Parkhi, A.~Vedaldi, and A.~Zisserman, ``Deep face recognition,'' in
  \emph{British Machine Vision Confererence}, 2015.

\bibitem{XWuT:2018}
X.~Wu, R.~He, Z.~Sun, and T.~Tan, ``A light {CNN} for deep face representation
  with noisy labels,'' \emph{IEEE Transactions on Information Forensics and
  Security}, vol.~13, no.~11, pp. 2884--2896, 2018.

\bibitem{Ronneberger:2015}
O.~Ronneberger, P.~Fischer, and T.~Brox, ``{U}-{N}et: Convolutional networks
  for biomedical image segmentation,'' in \emph{International Conference on
  Medical Image Computing \& Computer Assisted Intervention}, 2015, pp.
  234--241.

\bibitem{Kingma:2014}
D.~P. Kingma and J.~Ba, ``Adam: A method for stochastic optimization,'' in
  \emph{International Conference on Learning Representations}, 2015.

\bibitem{YDGuo:2106}
Y.~Guo, L.~Zhang, Y.~Hu, X.~He, and J.~Gao, ``Ms-celeb-1m: A dataset and
  benchmark for large-scale face recognition,'' in \emph{European Conference on
  Computer Vision (ECCV)}, 2016, pp. 87--102.

\bibitem{pix2pix2016}
P.~Isola, J.-Y. Zhu, T.~Zhou, and A.~A. Efros, ``Image-to-image translation
  with conditional adversarial networks,'' in \emph{IEEE Conference on Computer
  Vision and Pattern Recognition}, 2017.

\bibitem{YJin:2015}
Y.~Jin, J.~Lu, and Q.~Ruan, ``Coupled discriminative feature learning for
  heterogeneous face recognition,'' \emph{IEEE Transactions on Information
  Forensics and Security}, vol.~10, no.~3, pp. 640--652, 2015.

\bibitem{XLiu:2016}
X.~Liu, M.~Kan, W.~Wu, S.~Shan, and X.~Chen, ``Viplfacenet: An open source deep
  face recognition sdk,'' \emph{Frontiers of Computer Science}, 2016.

\bibitem{YWen:2016}
Y.~Wen, K.~Zhang, Z.~Li, and Y.~Qiao, ``A discriminative feature learning
  approach for deep face recognition,'' in \emph{European Conference on
  Computer Vision}, 2016.

\bibitem{ZLei:2009}
Z.~Lei and S.~Z. Li, ``Coupled spectral regression for matching heterogeneous
  faces,'' in \emph{IEEE Conference Computer Vision Pattern Recognition}, 2009,
  pp. 1123--1128.

\end{thebibliography}
}
\end{document}